\documentclass[conference]{IEEEtran}
\IEEEoverridecommandlockouts
\usepackage{cite}
\usepackage{amsmath,amssymb,amsfonts}
\usepackage{algorithmic}
\usepackage{graphicx}
\usepackage{textcomp}
\usepackage{xcolor}
\usepackage{multirow}

\usepackage{url}

\usepackage{hyperref}
\usepackage{booktabs}
\usepackage{subfig}
\usepackage{float}

\def\BibTeX{{\rm B\kern-.05em{\sc i\kern-.025em b}\kern-.08em
    T\kern-.1667em\lower.7ex\hbox{E}\kern-.125emX}}
\begin{document}

\title{ArabSign: A Multi-modality Dataset and Benchmark for Continuous Arabic Sign Language Recognition
}

\author{\IEEEauthorblockN{Hamzah Luqman}
\IEEEauthorblockA{Information and Computer Science Department, King Fahd University of Petroleum and Minerals\\ SDAIA-KFUPM Joint Research Center for Artificial Intelligence, Dhahran 31261, Saudi Arabia. \\
Email: hluqman@kfupm.edu.sa}}


\maketitle

\begin{abstract}

Sign language recognition has attracted the interest of researchers in recent years. 
While numerous approaches have been proposed for European and Asian sign languages recognition, very limited attempts have been made to develop similar systems for the Arabic sign language (ArSL). This can be attributed partly to the lack of a dataset at the sentence level.
In this paper, we aim to make a significant contribution by proposing ArabSign, a continuous ArSL dataset. The proposed dataset consists of 9,335 samples performed by 6 signers. The total time of the recorded sentences is around 10 hours and the average sentence's length is 3.1 signs. ArabSign dataset was recorded using a Kinect V2 camera that provides three types of information (color, depth, and skeleton joint points) recorded simultaneously for each sentence. In addition, we provide the annotation of the dataset according to ArSL and Arabic language structures that can help in studying the linguistic characteristics of ArSL. To benchmark this dataset, we propose an encoder-decoder model for Continuous ArSL recognition. The model has been evaluated on the proposed dataset, and the obtained results show that the encoder-decoder model outperformed the attention mechanism with an average word error rate (WER) of 0.50 compared with 0.62 with the attention mechanism. The data and code are available at \href{https://github.com/Hamzah-Luqman/ArabSign}{https://github.com/Hamzah-Luqman/ArabSign}

\end{abstract}

\section{Introduction}
\label{sec:introduction}

Hearing loss is a serious problem facing the world today, and it is getting worse. It is estimated that nearly 2.5 billion people are projected to have some degree of hearing loss by 2050, and at least 700 million will require hearing rehabilitation~\cite{whoOrganization}. Modern lifestyles and unsafe listening practices mean that over 1 billion young adults are at risk of permanent hearing loss.

Sign language is the main communication language of hearing impaired people. This language is a complete and rich language with grammar and structure that differ from spoken languages. Sign language has its own lexicon that is usually smaller than spoken languages' vocabulary.

\begin{figure}[!h]
    \centering
  \includegraphics[width=0.9\linewidth]{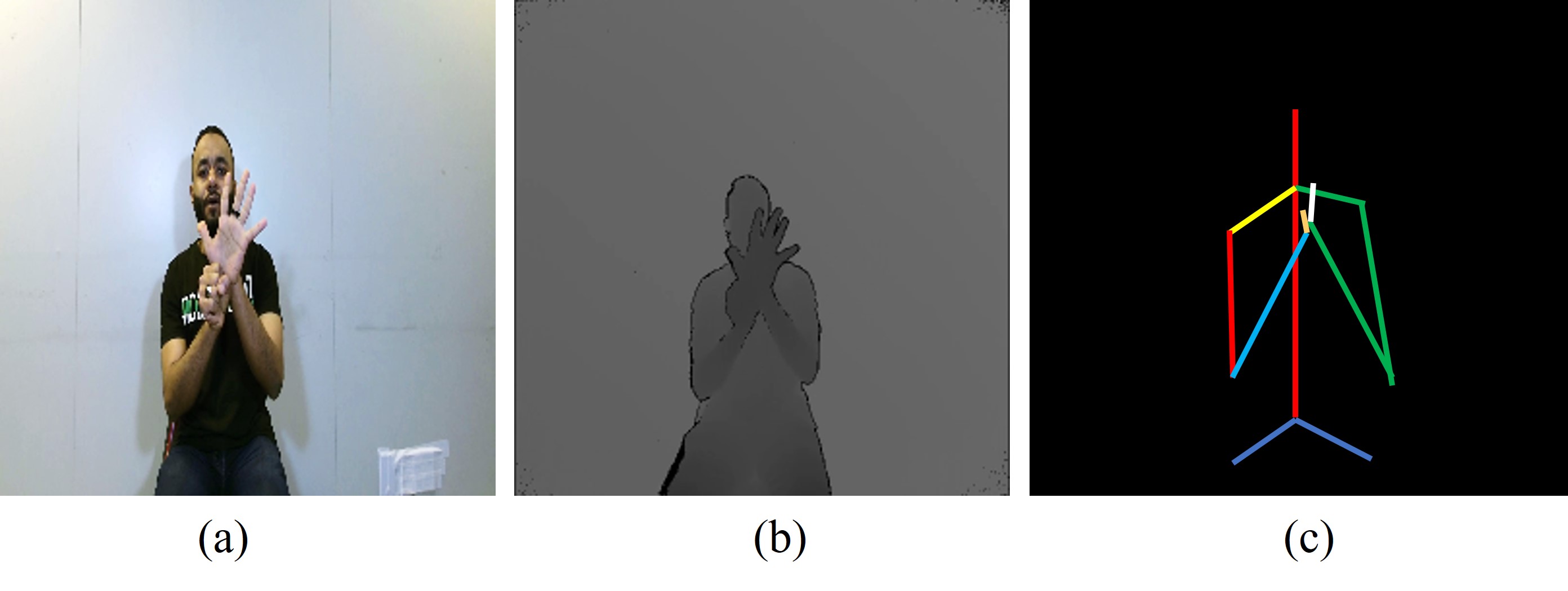}
    \caption{An illustrative example from the ArabSign dataset for the three modalities provided for each sentence sample: (a) color, (b) depth, and (c) skeleton joint points.  }
 \label{fig:kincet_output}
\end{figure}

Sign language is not a universal language and it does not depend on spoken languages~\cite{aloysius2020understanding}. Sign languages are "not mutually intelligible with each other" although there are some similarities in some signs. There are many sign languages that differ in their gestures, lexicon, and grammar. Most of the sign languages are related to the country more than the spoken language of that country. There are some countries that speak one language but have different sign languages, such as British Sign Language (BSL) and American Sign Language (ASL). Other popular sign languages are Chinese (CSL), German (GSL), Indian (ISL), and Arabic (ArSL) sign languages.
 ArSL is one of the main languages used in Arab countries. It is currently the main language used in translating television programs such as news and interviews.
This language has a dictionary consisting of 3200 sign words published in two parts~\cite{ArSLDict01, ArSLDict02}.

Sign language is a non-verbal language that uses multi-modality data to express thoughts~\cite{el2022comprehensive}. Manual and non-manual gestures are the two modalities used in sign language for communication. These gestures are combined during signing in a way that complements each other. Manual gestures are the dominant element used in sign languages. These gestures employ body movements through the hands and head. The majority of sign language signs depend on manual gestures. The non-manual modality consists mainly of facial expressions that are simultaneously performed with manual gestures. Non-manual gestures are used to show emotions and feelings in sign language in addition to linguistic properties such as grammatical structure, adjectival or adverbial content, and lexical distinction.

Translating sign language into spoken language is accomplished through sign language recognition (SLR) and translation~\cite{luqman2019automatic}. Automatic SLR involves using pattern recognition and computer vision to identify sign gestures and convert them into their equivalent words in the natural language~\cite{wadhawan2021sign}. Sign language translation involves using natural language processing and linguistics to translate the recognized sign language sentences into spoken languages to meet their structure and grammar. Extensive research has been conducted on SLR compared with translation since translation depends on the output of the SLR at the sentence level.

Based on the type of the recognized signs, SLR systems can be categorized into isolated and continuous SLR systems. Isolated sign recognition systems target isolated sign words while continuous sign language recognition (CSLR) systems target more than one sign performed continually.
Most of the techniques that have been proposed for SLR during the last three decades have targeted isolated signs \cite{el2022comprehensive}. CSLR is still in its infancy compared with isolated SLR, where the growth of CSLR studies is close to linear compared with the exponential growth of isolated SLR studies \cite{koller2020quantitative}.
One of the challenges associated with CSLR is the lack of movement epenthesis clues between sentence signs and the lack of temporal information that can help in signs segmentation. In addition, the high variance between signs performed by different signers made the learning of segmentation clues very difficult. Another challenge is the lack of datasets, which can be considered the main challenge that makes most of the researchers target isolated signs.

To our knowledge, no available vision-based annotated sentences of ArSL that can be used for ArSL CSLR and translation. The datasets that have been proposed for Arabic CSLR are collected using glove sensors. However, sensor-based SLR requires signers to keep wearing the electronic sensor gloves during signing. This makes these sensors unsuitable for real-time applications. In addition, the sensors used for sign acquisition can not capture the non-manual features of the sign language. This motivated us to propose a continuous ArSL dataset that can be used for CSLR and translation.
The main contributions of this research are as follows:
\begin{itemize}
\item Propose a continuous ArSL dataset (ArabSign). The proposed dataset was collected using a multi-modality Kinect V2 camera. The dataset is available in three modalities: color, depth, and joint points shown in Figure~\ref{fig:kincet_output}. The proposed dataset consists of 9,335 samples representing 50 ArSL sentences. Each sentence was performed by 6 signers, and each sentence was repeated several times by each signer.
\item Provide the annotation of the performed sentences according to the structure of ArSL and Arabic language. This makes the dataset useful for studying the grammar and structure of ArSL and developing machine translation systems between ArSL and natural languages.
\item Propose an encoder-decoder model for benchmarking the proposed ArabSign dataset. The model has been trained on features extracted from the color frames of the sentences using different pre-trained models. In addition, the proposed model has been compared with an attention mechanism.

\end{itemize}



This paper is organized as follows: a literature review of the available continuous sign language datasets is presented in Section \ref{sec:literature_review}. A detailed description of the proposed ArabSign dataset is presented in Section \ref{sec:dataset}. Section \ref{sec:system_arch} describes the experimental work that has been conducted to benchmark the proposed dataset, and the conclusions are presented in Section \ref{sec:conclusions}.

\section{Literature review}
\label{sec:literature_review}

The work on SLR can be dated back to the middle of the 1990s \cite{koller2020quantitative}. 
The SLR systems at the sign level are the most common SRL systems compared with the sentence level due to the availability of datasets at the sign level and the similarity between this problem and gesture recognition problems \cite{el2022comprehensive}. In contrast, few approaches have been proposed for CSLR due to the challenges associated with recognizing sign languages' sentences. 
One of these challenges is the lack of annotated datasets.

Few datasets have been proposed for continuous SL compared with isolated sign datasets. The majority of these datasets target ASL and DGS. There are some datasets that are used by researchers for their work. However, these datasets are either limited in size or unavailable for researchers. The most commonly used continuous sign language datasets were proposed by a group at RWTH Aachen University. This group proposed four datasets for continuous ASL and DGS, namely RWTH-BOSTON-104~\cite{dreuw2007speech}, RWTH-BOSTON-400~\cite{dreuw2008benchmark}, RWTH-PHOENIX-Weather~\cite{forster2012rwth}, and RWTH-PHOENIX-Weather-2014~\cite{forster2014extensions}. RWTH-BOSTON-104 \cite{dreuw2007speech} was recorded at Boston University and it consists of 201 sentences of ASL performed by three signers. The vocabulary size of this dataset is 168 sign words.

RWTH-BOSTON-400~\cite{dreuw2008benchmark} is an extension of RWTH-BOSTON-104. It consists of 843 sentences with a vocabulary size of 406 sign words performed by four signers. RWTH-PHOENIX-Weather~\cite{forster2012rwth} includes weather forecasts collected from German television. This dataset is performed by seven signers and it consists of 1,980 sentences of DGS with a vocabulary size of 911 sign words. This dataset is extended in RWTH-PHOENIX-Weather-2014 \cite{forster2014extensions} to 6,861 sentences performed by nine signers. Both datasets were recorded in a controlled environment where signers were wearing a dark T-shirt with grey background. How2Sign~\cite{Duarte_2021_CVPR} is a multi-view ASL dataset consisting of around 35K samples performed by 11 signers for a duration of 79 hours.

SIGNUM \cite{von2008significance} is a DGS dataset consisting of 780 sentences performed by 25 signers. The SignsWorld Atlas \cite{shohieb2015signsworld} is an ArSL dataset consisting of five sentences performed by four signers. TheRuSLan \cite{kagirov2020theruslan} is a Russian SL dataset consisting of 164 sentences performed by 13 signers. Huang et al. \cite{huang2018video} proposed a CSL dataset consisting of 100 sentences with a vocabulary size of 178 sign words performed by 50 signers. Table \ref{table_db_lr} summarizes the available continuous sign language datasets. The missing information in the table is not reported in the respective reference. 



\begin{table*}[]
 \caption{A summary of the publicly available continuous sign language datasets.}
\centering
\small 
   \label{table_db_lr}
\begin{tabular}{lcccccc} 
\toprule
\textbf{Dataset}          & \textbf{Language} & \textbf{Sentences} & \textbf{Duration (h)} & \textbf{Vocabulary Size} & \textbf{Signers} & \textbf{Samples}  \\ 
\midrule
RWTH-BOSTON-104 \cite{dreuw2007speech}          & ASL               &       400   &  -   &      104                 & 3      &    -     \\ 
RWTH-BOSTON-400 \cite{dreuw2008benchmark}          & ASL               & 843    &    -        & 400                      & 4        &  -       \\ 

MS-ASL \cite{joze2018ms} & ASL & 1000 & - & - & 222 & 25,513 \\ 
How2Sign \cite{Duarte_2021_CVPR} & ASL & - & 79 & 16K & 11 & 35,191 \\

RWTH-PHOENIX-Weather \cite{forster2012rwth}     & DGS               & 1,980     &     -     & 911                      & 7           &   1,980   \\ 

RWTH-PHOENIX-Weather-2014 \cite{forster2014extensions} & DGS               & 6,861    &    -       & 1,080                    & 9    &      6,861       \\ 

SIGNUM  \cite{von2008significance}                  & DGS               & 780   &    55          & 455                      & 25    &     780       \\ 

TheRuSLan  \cite{kagirov2020theruslan}               & Russian           & 164  &      -         &      -                   & 13    &   -         \\ 

Video-based CSL \cite{huang2018video}               & CSL               & 100    &   100           & 178                      & 50      &  5,000        \\

SignsWorld Atlas \cite{shohieb2015signsworld}         & ArSL              & 5     &             &    -                     & 4       &    -      \\ 
\bottomrule
\end{tabular}
\end{table*}

Another challenge of CSLR is sign segmentation. This can be attributed to the lack of movement epenthesis clues between sentence's signs and the lack of temporal information that can help in sign segmentation \cite{el2022comprehensive}. In addition, the high variance between signs performed by different signers made the learning of segmentation clues very difficult. These challenges motivated researchers to use recognition techniques that do not require pre-segmented sentences such as Hidden Markov models \cite{brashear2003using, mcguire2004towards, Mohandes2013, zhang2015new, nagendraswamy2017lbpv, hassan2019multiple} and deep learning techniques \cite{ camgoz2017subunets, huang2018video, cui2019deep, pu2019iterative ,zhou2019dynamic, papastratis2020continuous, tamer2020keyword, zhou2020spatial, luqman2021towards }. However, some researchers converted the CSLR problem into an isolated signs recognition problem by segmenting the sentences into signs and recognizing each sign separately \cite{gao2004transition, zhang2014threshold, zhang2015new, kong2014towards,  farag2019learning, sidig2019arabic, ye2018recognizing }.

\section{ArabSign dataset}
\label{sec:dataset}

\begin{figure}
    \centering
  \includegraphics[width=0.8\linewidth]{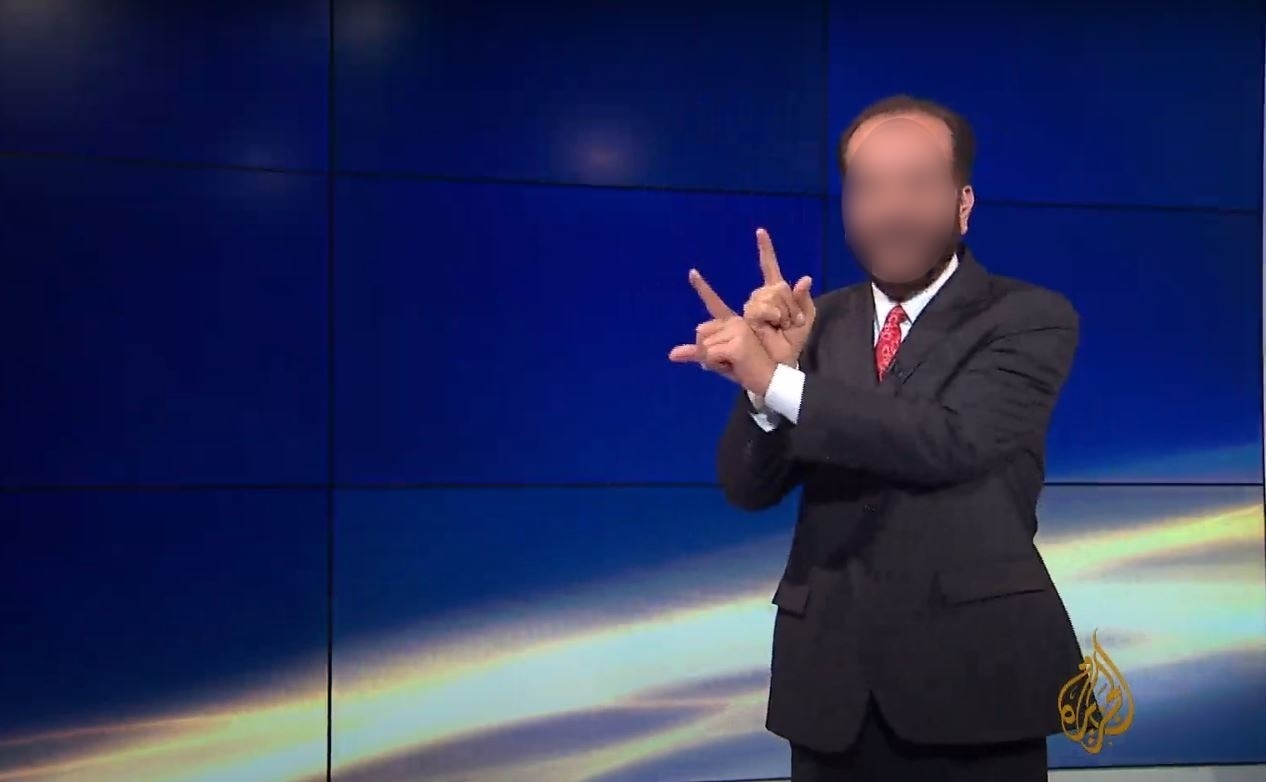}
    \caption{A sample of Al-Jazeera ArSL sentence videos that was followed by the signers of ArabSign dataset~\cite{AljazeeraSL}.}
 \label{fig:dataset_reference}
\end{figure}

The ArabSign dataset consists of ArSL video sentences with their corresponding annotations in Arabic and English languages. A total of 10 hours and 13 minutes of ArSL were recorded by 6 signers in different recording sessions. The sentences used in this dataset are based on the ArSL tutorial videos performed by ArSL translation experts and produced by Al-Jazeera media network~\cite{AljazeeraSL}.
We extracted fifty most commonly used sentences in ArSL from these videos. Then, we used these sentences as a reference for our signers, and each signer was asked to repeat those sentences. The reference sentences will be published with the dataset for interested researchers. A sample of a reference video is shown in Figure \ref{fig:dataset_reference}.

\begin{figure*}
    \centering
  \includegraphics[width=\linewidth]{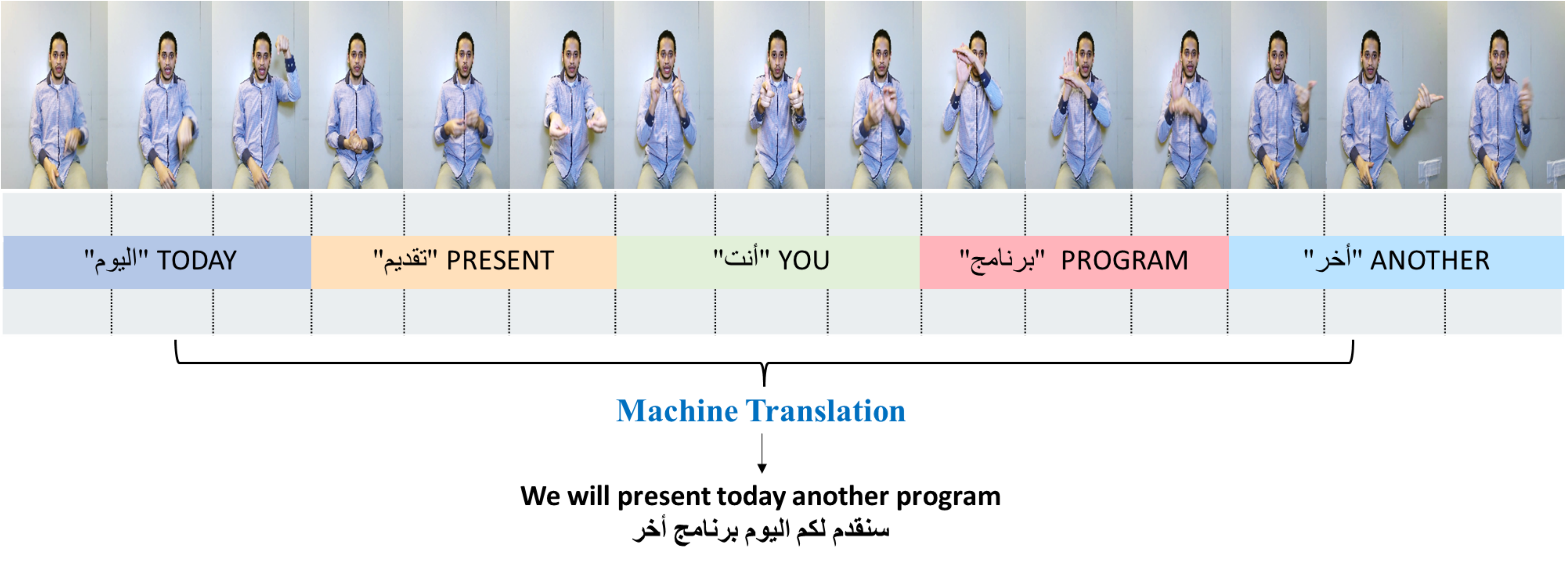}
    \caption{ArabSign sentence with its glosses and the corresponding sentence resulting from machine translation systems. }
 \label{fig:dataset_annotation}
\end{figure*}

All the sentences of the proposed dataset were annotated with their glosses in both Arabic and English languages. The annotation followed the structure of the ArSL sentence as performed by the signer. Figure \ref{fig:dataset_annotation} shows an example of an ArabSign sentence with its glosses and its equivalent in Arabic language. To make the dataset useful for machine translation tasks between ArSL and Arabic language, we also provide the annotation of each sentence according to the structure and grammar of Arabic language. This helps in studying the linguistic characteristics of ArSL.

\subsection{Recording setup}
The ArabSign dataset was performed by 6 signers. Three of these signers have experience with sign language and have been involved in sign language gesturing before. The other signers were trained on sign language during the project. The signers first watched the sentences video performed by an expert in sign language translation who is working in Al-Jazeera news media. The video was played at a slow speed of 0.75 to help the signers in learning the sentence's signs correctly.
Each sentence was transcripted to show the sign of each word to help the signers be familiar with the sentence glosses. The signers were asked to repeat each ArSL sentence several times before the recording session to minimize the variations between sentence samples.

The dataset's sentences have been recorded in several sessions. Each signer was asked to wear casual clothes and use different clothes in each recording session to add more variability to the dataset. The dataset was recorded in an unconstrained environment. We used the room lights during the recordings. The sessions were recorded at different times of the day. In addition, all the sentences were recorded with a white background.

The dataset was recorded using Microsoft Kinect V2. This camera is a multi-modality camera that provides three types of information recorded simultaneously for each sentence. It provides color, depth, and skeleton information. The color video is recorded with a resolution of 1920 $\times$ 1080 at 30 fps. The depth information is available as a video stream with a resolution of 512 $\times$ 424. Kinect V2 provides 25 joint points for each signer. The joint points were captured for each frame and are available in Matlab file format. This file contains the position of each joint point in the camera space, represented using X, Y, and Z coordinates. It also contains the orientation, tracking state, left and right hands' states, and the cardinal position of each joint point in the color and depth videos. 

\subsection{Dataset statistics}
The dataset consists of 9,335 samples representing 50 sentences of ArSL. The dataset's sentences were performed by 6 signers. Each sentence was repeated by each signer at least 30 times at different sessions. All signers are male with different skin colors. The signers' ages range between 21 and 30 years old. All signers are right-handed, and one of them was wearing eyeglasses.

\begin{figure}[]
 \centering

\includegraphics[width=0.4\textwidth]{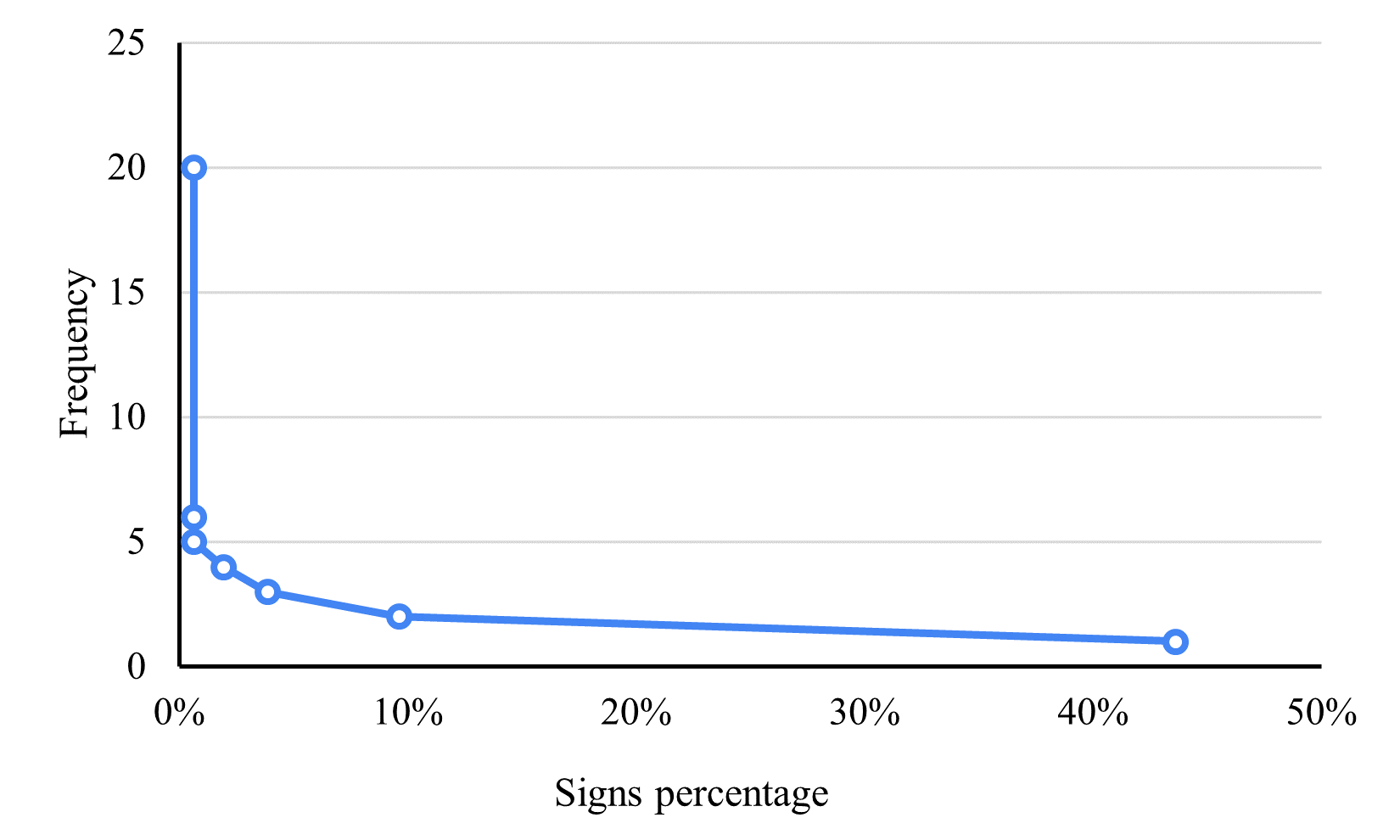}

    \caption{Signs' frequency in the dataset. }
 \label{fig:signs_count}
\end{figure}

The dataset's sentences consist of 155 signs, and the dataset's vocabulary consists of 95 signs. Figure \ref{fig:signs_count} shows the frequency of the signs in the dataset. As shown in the figure, more than 40\% of the dataset signs appeared less than 5 times. Having a large number of unique signs or signs that appear a few times makes the dataset appropriate for evaluating real-time recognition systems.

\begin{figure}[]
 \centering
\includegraphics[width=0.45\textwidth]{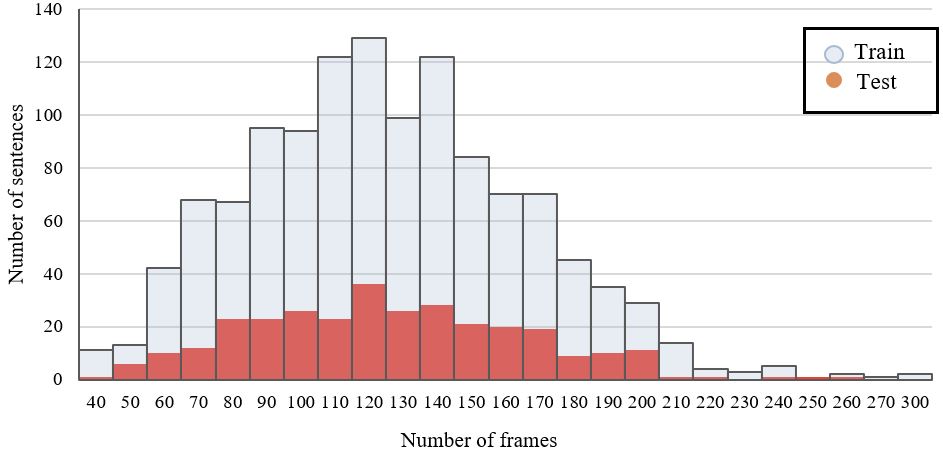}
    \caption{Frames' frequency over sentences.  }
 \label{fig:frames_count}
\end{figure}

The total time of the recorded sentences is around 10 hours and 13 minutes. The duration of each sentence depends on its length in terms of the number of signs and the signer's signing speed. The average sentence length is 3.1 signs. The dataset was recorded at the normal speed of signing. Each sentence was signed continuously with no pauses between sentence's signs. This resulted in around 200,000 frames for all sentences performed by one signer, with an average of 130.3 frames per sentence, as shown in Figure \ref{fig:frames_count}. This figure shows the frames over sentence level clips for one signer. Table \ref{table:db_statistcs} shows the statistics of the proposed ArabSign dataset.


\begin{table}
\centering
\caption{Statistics of the proposed ArabSign dataset}
\label{table:db_statistcs}
\resizebox{\linewidth}{!}{%
\begin{tabular}{ll|ll}
\hline
RGB resolution             & 1920×1080 & \# of signers        &    6    \\ \hline
Depth resolution           & 512×424   & Vocab. size          &   95     \\ \hline
Body joints                & 21        & Average words/sample &    3.1    \\ \hline
Min. video duration (sec.) &      1.3     & Repetitions/sentence &  $\geq$30      \\ \hline
Max. video duration (sec.) &   10.4        & FPS                  & 30     \\ \hline
Total hours                & 10.13     & Total Samples        & 9,335 \\ \hline
\end{tabular}}
\end{table}


\section{Experimental Evaluation}
\label{sec:system_arch}

\subsection{System Architecture}
\begin{figure*}[h]
 \centering
\includegraphics[width=\textwidth]{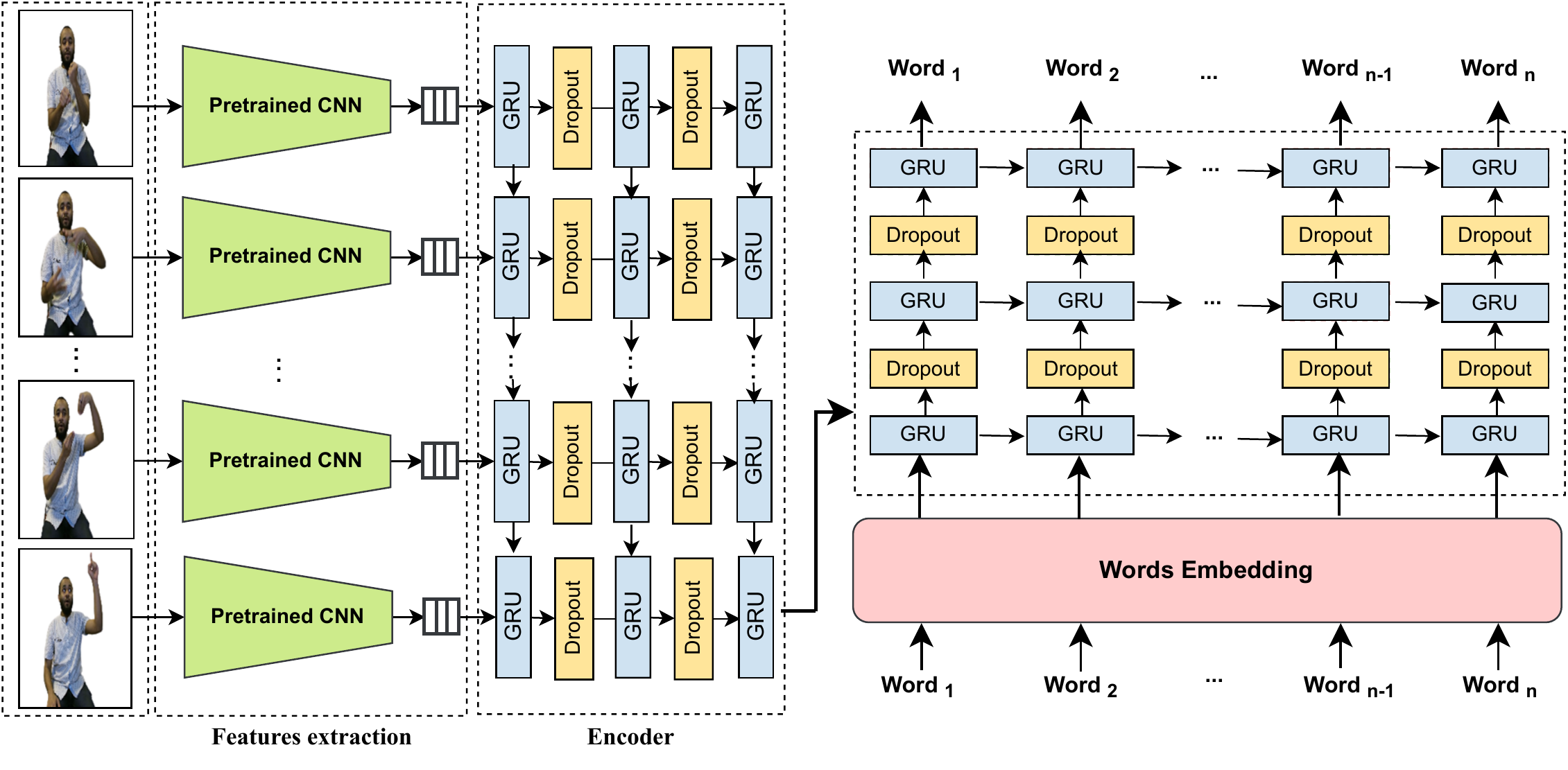}
    \caption{The framework of the encoder-decoder model.  }
 \label{fig:framework}
\end{figure*}

In this work, we propose a continuous ArSL recognition framework for benckmarking the ArabSign dataset. The architecture of the proposed framework is shown in Figure~\ref{fig:framework}. The features are extracted from sentence videos using pre-trained models and fed into an encoder-decoder model for features learning and classification. To learn the corresponding natural language text of the recognized sentences, we employed word embedding. This section describes the components of the proposed framework for CSLR.

\subsubsection{Features Extraction}
\label{sect:featuresExtraction}
Sign language learning depends on two types of information, spatial and temporal. Spatial information in sign language represents the shape and orientation of the body parts used during signing, such as the hand, head, and mouth. The temporal information of sign gestures involves the motion of the signer's body parts during signing.  

Spatial information is important for sign language understanding. Learning these features efficiently contributes to improving the model accuracy. Although time-series learning techniques, such as Long short-term memory (LSTM) and Gated recurrent unit (GRU), are efficient for temporal information learning, they lack the ability to learn spatial information \cite{wang2018human}. To address this issue, we employed CNN models for features extraction.

Two pre-trained CNN models are used in this work for extracting the spatial features from the sign frames. These models are MobileNet~\cite{howard2017mobilenets} and InceptionV3~\cite{szegedy2016rethinking}. All of these models have been trained originally on the ImageNet dataset, which consists of images grouped into 21,841 subcategories. Although these models have been trained on ImageNet, the performance of each model can vary due to the structure and specifications of the model. These differences make each model appropriate for different computer vision tasks.

Different number of features were extracted from each model. For each frame in the sign gesture, 1024 features were extracted using the MobileNet pre-trained model. We also extracted 2048 features from each frame using the InceptionV3 model. The variation in the number of extracted features is related to the architecture of each model. These features are used as inputs to the proposed models.

\subsubsection{Words Embedding}

The CSLR system accepts the sentence gestures as a sequence of frames and outputs their equivalent glosses in the form of an ArSL sentence. The sentence frames are fed into the encoder, and the ground truth of the sentence will be fed into the decoder during the training stage of the proposed models. The ground truth is an Arabic sentence consisting of a sequence of glosses representing the signs of the sentence, as shown in Figure \ref{fig:dataset_annotation}.

Several techniques have been proposed in the literature for word representation, such as TF-IDF and N-gram. These techniques ignore the word context that can affect the performance of several natural language processing systems. Therefore, word embedding techniques have been proposed recently to address this issue~\cite{Yadav2020}. These embeddings played an important role in boosting the performance of several machine learning models~\cite{Devlin2019}. In this work, we used an embedding layer to represent each sentence's word as a vector of size 300 and used these vectors as an input to the decoder.


\begin{table*}[!ht]
\centering
\caption{The performance of the proposed models in the signer-dependent mode. }
\label{table:signer_depedent_wer}
\resizebox{0.9\textwidth}{!}{
\begin{tabular}{l||cccccccc}
\toprule
          \multirow{2}{*}{}        & \multicolumn{2}{c}{\textbf{Encoder-decoder-Inception}} & \multicolumn{2}{c}{\textbf{Encoder-decoder-MobileNet}} & \multicolumn{2}{c}{\textbf{Attention-Inception}}   & \multicolumn{2}{c}{\textbf{Attention-MobileNet}}   \\   \cmidrule{2-3}  \cmidrule{4-5}  \cmidrule{6-7} \cmidrule{8-9}
                  & \multicolumn{1}{c}{\textrm{BLEU-4}}   & WER   & \multicolumn{1}{c}{\textrm{BLEU-4}}   & WER   & \multicolumn{1}{c}{\textrm{BLEU-4}} & WER & \multicolumn{1}{c}{\textrm{BLEU-4}} & WER \\   \cmidrule{1-9}
Signer 1          & \multicolumn{1}{c}{0.33}              & 0.00           & \multicolumn{1}{c}{0.32}              & 0.01           & \multicolumn{1}{c}{0.33}            & 0.01         & \multicolumn{1}{c}{0.32}            & 0.02         \\ 
Signer 2          & \multicolumn{1}{c}{0.32}              & 0.05           & \multicolumn{1}{c}{0.32}              & 0.02           & \multicolumn{1}{c}{0.32}            & 0.01         & \multicolumn{1}{c}{0.32}            & 0.00         \\ 
Signer 3          & \multicolumn{1}{c}{0.33}              & 0.00           & \multicolumn{1}{c}{0.33}              & 0.00           & \multicolumn{1}{c}{0.33}            & 0.00         & \multicolumn{1}{c}{0.33}            & 0.00         \\ 
Signer 4          & \multicolumn{1}{c}{0.34}              & 0.02           & \multicolumn{1}{c}{0.34}              & 0.00           & \multicolumn{1}{c}{0.34}            & 0.01         & \multicolumn{1}{c}{0.34}                & 0.01         \\ 
Signer 5          & \multicolumn{1}{c}{0.32}              & 0.01           & \multicolumn{1}{c}{0.32}              & 0.01           & \multicolumn{1}{c}{0.32}            & 0.00         & \multicolumn{1}{c}{0.32}                & 0.00         \\ 
Signer 6          & \multicolumn{1}{c}{0.32}              & 0.06           & \multicolumn{1}{c}{0.32}              & 0.02           & \multicolumn{1}{c}{0.32}            & 0.01         & \multicolumn{1}{c}{0.31}                & 0.11         \\ 
All               & \multicolumn{1}{c}{0.32}              & 0.11           & \multicolumn{1}{c}{0.30}              & 0.14           & \multicolumn{1}{c}{0.31}            & 0.06         & \multicolumn{1}{c}{0.33}                & 0.04         \\ \midrule
\textbf{Average}           & \multicolumn{1}{c}{0.33}              & 0.04           & \multicolumn{1}{c}{0.32}              & 0.03           & \multicolumn{1}{c}{0.32}           & 0.01        & \multicolumn{1}{c}{0.32}               & 0.03         \\ \bottomrule
\end{tabular}
}
\end{table*}


\subsubsection{Encoder-Decoder model}

An encoder-decoder model has been proposed in this work. The structure of the proposed model is shown in Figure~\ref{fig:framework}. The encoder component accepts the features extracted from each sign's frame using the pre-trained models discussed in Section \ref{sect:featuresExtraction}. Each feature vector is fed into the encoder component, which consists of three bidirectional GRU layers separated by a dropout layer. Each GRU layer consists of 1024 neurons. To mitigate the overfitting, we used two dropout layers with a probability of 0.4. These hyper-parameters have been selected empirically. 

The decoder component of the model is responsible for generating the equivalent sentence of the sign sentence video. This model accepts two inputs during training. The first input is the output of the encoder component, and the second input is the sentence's ground truth. The ground truth input is represented as a sequence of word glosses. These words are fed first into a word embedding layer and the output of this layer is used as an input to the decoder, as illustrated in Figure~\ref{fig:framework}.
The decoder model consists of three stacked GRU layers with 1024 neurons, selected empirically. The outputs of the first two GRU layers are fed into dropout layers with a probability of 0.4. The output of the model is a sequence of word glosses representing the recognized sentence.
   
We also proposed a variant of the encoder-decoder model with attention layer. We will refer to this variant by attention model.
The attention mechanism is a deep learning technique that provides more focus on some components of the input. Attention was proposed first for neural machine translation~\cite{bahdanau2014neural}. Later, this mechanism, or its variants, was used in other applications, including speech processing, computer vision, etc.

The attention model consists of four components. The convolutional component which is used to extract features from each sign frame, as discussed in Section \ref{sect:featuresExtraction}. The feature vector of each frame is fed into the encoder component, which consists of three stacked bidirectional GRU layers separated by two dropout layers. The architecture of the encoder component is similar to the encoder-decoder model discussed before. 

The output of the encoder component is fed into the decoder and attention fusion components. We used the attention fusion module as a layer to obtain the attention weights, which were multiplied by the encoder’s output to obtain the attended encoder output. We use this attended encoder output as a context tensor, which represents a weighted sum indicating which parts of the encoder’s output to pay attention to. The output of the decoder with attention is used to predict the corresponding word gloss.

\begin{table*}[!ht]
\centering
\caption{The performance of the proposed models in the signer-independent mode. }
\label{table:signer_indepedent_wer}
\resizebox{0.9\textwidth}{!}{
\begin{tabular}{l||cccccccc}
\toprule
          \multirow{2}{*}{}        & \multicolumn{2}{c}{\textbf{Encoder-decoder-Inception}} & \multicolumn{2}{c}{\textbf{Encoder-decoder-MobileNet}} & \multicolumn{2}{c}{\textbf{Attention-Inception}}   & \multicolumn{2}{c}{\textbf{Attention-MobileNet}}   \\   \cmidrule{2-3}  \cmidrule{4-5}  \cmidrule{6-7} \cmidrule{8-9}
                  & \multicolumn{1}{c}{\textrm{BLEU-4}}   & WER   & \multicolumn{1}{c}{\textrm{BLEU-4}}   & WER   & \multicolumn{1}{c}{\textrm{BLEU-4}} & WER & \multicolumn{1}{c}{\textrm{BLEU-4}} & WER \\   \cmidrule{1-9}
Signer 1 & \multicolumn{1}{c}{0.20}               & 0.44           & \multicolumn{1}{c}{0.27}              & 0.27           & \multicolumn{1}{c}{0.18}            & 0.54         & \multicolumn{1}{c}{0.24}            & 0.34         \\
Signer 2 & \multicolumn{1}{c}{0.24}              & 0.46           & \multicolumn{1}{c}{0.19}              & 0.51           & \multicolumn{1}{c}{0.21}            & 0.61         & \multicolumn{1}{c}{0.16}            & 0.65         \\ 
Signer 3 & \multicolumn{1}{c}{0.19}              & 0.50            & \multicolumn{1}{c}{0.19}              & 0.6            & \multicolumn{1}{c}{0.15}            & 0.64         & \multicolumn{1}{c}{0.16}            & 0.76         \\ 
Signer 4 & \multicolumn{1}{c}{0.13}              & 0.67           & \multicolumn{1}{c}{0.14}              & 0.58           & \multicolumn{1}{c}{0.09}            & 0.76         & \multicolumn{1}{c}{0.14}            & 0.52         \\ 
Signer 5 & \multicolumn{1}{c}{0.25}              & 0.36           & \multicolumn{1}{c}{0.22}              & 0.37           & \multicolumn{1}{c}{0.18}            & 0.53         & \multicolumn{1}{c}{0.12}            & 0.67         \\ 
Signer 6 & \multicolumn{1}{c}{0.12}              & 0.61           & \multicolumn{1}{c}{0.06}              & 0.65           & \multicolumn{1}{c}{0.09}            & 0.66         & \multicolumn{1}{c}{0.04}            & 0.76         \\ \midrule
\textbf{Average}  & \multicolumn{1}{c}{0.19}              & 0.51           & \multicolumn{1}{c}{0.18}              & 0.50            & \multicolumn{1}{c}{0.15}            & 0.62         & \multicolumn{1}{c}{0.14}            & 0.62         \\\bottomrule
\end{tabular}
}
\end{table*}

\subsection{Results and Discussion}
\label{sec:experimts}
Several experiments have been conducted to evaluate the proposed models and benchmark the proposed ArabSign dataset.
We ran all the experiments using Pytorch 1.12 on a workstation with an Nvidia GeForce RTX 2080 TI GPU with 11 GB of GPU memory and 64 GB of RAM memory.

We evaluated the proposed models in the signer-dependent and signer-independent modes. In signer-dependent mode, the models are trained and tested on samples from the same signer(s). In contrast, the signer-independent mode involves testing the model on a signer that has not been seen during the model training.

We used BLEU and WER metrics to evaluate the proposed models. Bilingual Evaluation Understudy (BLEU) performs n-gram matching between the sentences resulting from the CSLR models and the reference sentences \cite{papineni2002bleu}. We used the BLEU metric with a 4-gram and a brevity penalty in all our experiments.
The WER is derived from the Levenshtein distance and it works at the word level by comparing the CSLR output and the reference sentence word by word. This metric finds the differences between these sentences by computing the number of insertions, deletions, and substitutions normalized by the total number of words in the sentence. Both BLEU and WER metrics score ranges between 0 and 1, where 1 indicates an exact match between the CSLR output and the reference sentences.

Table \ref{table:signer_depedent_wer} shows the obtained BLEU-4 and WER results in the signer-dependent mode. We trained and tested the models on samples of one signer. We refer to these settings in the table by Signer 1, Signer 2..etc. We also combined all signers' samples and split them into training and testing, and we refer to this setting as 'All'. As shown in the table, the proposed models performed well with all signers' data in the signer-dependent mode. The lowest WER was obtained using an attention model with an Inception pre-trained network. In addition, the BLEU score was almost similar across all models. It is also noticeable that all models were able to recognize the sentences of signer 3 with a WER of 0, whereas signer 6 was the most challenging signer for all models. This can be attributed to the small variations between the samples of signer 3 in contrast to signer 6, who is not an expert in sign language. This resulted in some variations between the samples of the same sign. These variations affected the capabilities of the models in sentence learning and recognition.

CSLR in the signer-independent mode is more challenging than recognition in the signer-dependent mode, as can be seen from the reported results in Table \ref{table:signer_indepedent_wer}. As shown in the table, the obtained results using the encoder-decoder models outperformed the attention models with all signers. In addition, the lowest WER is obtained using the encoder-decoder model trained with the MobileNet pre-trained model. It is also noticeable that the highest WER is obtained with signer 6. 
These results align with the obtained results with signer 6 in the signer-dependent mode that reveal the variations between the signer's sentences.

\section{Conclusions}
\label{sec:conclusions}
CSLR is a hot computer vision problem with several approaches that have been proposed in the literature for CSLR. Most of these techniques depend on signs segmentation, whereas a few techniques recognize the sentence without the need for sign segmentation. To our knowledge, no technique has been proposed for continuous ArSL recognition. This can be attributed to the lack of an ArSL dataset at the sentence level. In this work, we proposed a continuous ArSL dataset. The dataset is composed of 9,335 samples, performed by 6 signers. The dataset has been acquired using Kinect V2 and all the samples are available in three forms: color, depth, and joint points. The dataset will be made publicly available to researchers.

Moreover, we have proposed encoder-decoder and attentions models for CSLR. The spatial features have been extracted from the sentence frames using two pre-trained models that are fed into the proposed models. We evaluated the models on the proposed dataset in the signer-dependent and independent modes. The obtained results show that the encoder-decoder model with features extracted using the MobileNet pre-trained network outperformed other models in the signer-independent mode.

\section*{Acknowledgment}

The author would like to acknowledge the support received from Saudi Data and AI Authority (SDAIA) and King Fahd University of Petroleum and Minerals (KFUPM) under SDAIA-KFUPM Joint Research Center for Artificial Intelligence Grant JRC-AI-RFP-05.

\end{document}